\documentclass[letterpaper, 10 pt, conference]{ieeeconf}  

\IEEEoverridecommandlockouts                              

\usepackage{graphicx}      
\usepackage{booktabs}
\usepackage{subcaption}
\usepackage{amsmath}
\usepackage{amssymb}
\usepackage{algorithm}
\usepackage{color}
\usepackage{algorithmic}
\usepackage[acronym]{glossaries}
\makeglossaries

\newacronym{DFJSP}{DFJSP}{Dynamic and Flexible Job Shop Scheduling}

\newacronym{FJSPs}{FJSPs}{Flexible Job Shop Scheduling Problems}
\newacronym{JSPs}{JSPs}{Job Shop Scheduling Problems}

\newacronym{SPT}{SPT}{Shortest Processing Time}
\newacronym{LPT}{LPT}{Longest Processing Time}
\newacronym{FIFO}{FIFO}{First In First Out}
\newacronym{LIFO}{LIFO}{Last In First Out}
\newacronym{MWR}{MWR}{Most Work Remaining}

\newacronym{MIN}{MIN}{Minimum Processing Time}
\newacronym{MINC}{MINC}{Minimum Completion Time}

\newacronym{NSGA-II}{NSGA-II}{Non-dominated Sorting Genetic Algorithm II}

\newacronym{NRGA}{NRGA}{Non-dominated Ranked Genetic Algorithm}

\newacronym{GAE}{GAE}{Generalized Advantage Estimation}

\newacronym{DRL}{DRL}{Deep Reinforcement Learning}

\newacronym{RL}{RL}{Reinforcement Learning}

\newacronym{POMDP}{POMDP}{Partially Observable Markov Decision Process}

\newacronym{MDP}{MDP}{Markov Decision Process}

\newacronym{PPO}{PPO}{Proximal Policy Optimization}

\newacronym{SB3}{SB3}{Stable-Baselines3}

\newacronym{RUL}{RUL}{Remaining Useful Life}

\newacronym{MILP}{MILP}{Mixed Integer Linear Programming}
\newacronym{AT-MILP}{AT-MILP}{Arrival Triggered Mixed Integer Linear Programming}
\newacronym{Best HH}{Best HH}{Best Heuristics in Hindsight}

\newacronym{MLE}{MLE}{Maximum Likelihood Estimate}

\newacronym{MLP}{MLP}{Multi-Layer Perceptron}

\newacronym{MAP}{MAP}{Maximum A Posteriori}

\title{\LARGE \bf
Deep Reinforcement Learning for Flexible Job Shop Scheduling with Random Job Arrivals
}

\author{
Yu Tang$^{1}$,
Muhammad Zakwan$^{2}$,
Efe C. Balta$^{2}$,
John Lygeros$^{3}$,
Alisa Rupenyan$^{1}$%
\thanks{$^{1}$ Centre for Artificial Intelligence, Zurich University of Applied Sciences, Switzerland}
\thanks{$^{2}$ Inspire AG, Zurich, Switzerland}
\thanks{$^{3}$ Automatic Control Laboratory, ETH Zurich, Switzerland}
}

\begin{document}

\maketitle
\thispagestyle{empty}
\pagestyle{empty}

\begin{abstract}


The Flexible Job Shop Scheduling Problem (FJSP) is the optimal allocation of a set of jobs to machines.
Two primary challenges persist in FJSP: the unpredictable arrival of future jobs and the combinatorial complexity of the problem, rendering it intractable for conventional mixed-integer linear programming solvers.
This paper proposes an event-based \gls{DRL} approach to solve FJSP with random job arrivals. Specifically, we employ the Proximal Policy Optimization algorithm and use lightweight Multi-Layer Perceptrons to train the \gls{DRL} agent for minimizing the total completion time of all jobs. We design the state representation to be directly accessible from the environment, and limit the learning agent to selecting from among a set of well-established dispatching rules. Simulations show that our \gls{DRL} approach outperforms any of the individual dispatching rules on datasets with varying heterogeneity and job arrival rates. We benchmark our \gls{DRL} against an arrival-triggered mixed-integer linear programming solution and show that our method achieves good performance especially when the datasets are heterogeneous.



\end{abstract}

\section{Introduction}
Personalized orders and mass customization have increased the importance of scheduling problems such as the \gls{JSPs}, which focus on allocating job orders to machines. In the \gls{JSPs}, each job consists of a sequence of operations that must follow predefined precedence constraints. The \gls{FJSPs} generalize the classical \gls{JSPs} by allowing each operation to be processed on one of several compatible machines, possibly with different processing times.
This additional flexibility significantly enlarges the solution space, making \gls{FJSPs} NP-hard optimization problems \cite{lv2025deep}. 

Static \gls{FJSPs} assume full prior knowledge of shop floor information and a deterministic environment. They can be formulated as \gls{MILP} \cite{manne1960job} or constraint programming problems \cite{baptiste2001constraint}, which can be solved using modern optimization solvers such as Gurobi \cite{gurobi} or OR-Tools CP-SAT \cite{cp_sat_google_report}, respectively. However, to address dynamic and uncertain conditions in real production environments, scheduling schemes must be continuously adapted \cite{wang2021adaptive}. Here, we aim to minimize the makespan of \gls{FJSPs} with random job arrivals, by modeling the problem as a \gls{MDP} \cite{puterman2014markov} and constructing an event-based Deep Reinforcement Learning (\gls{DRL}) environment. A standard \gls{PPO} algorithm \cite{schulman2017ppo} is trained using lightweight \gls{MLP} for both the actor and critic networks, and the \gls{DRL}  agent learns to make sequential actions by implicitly assigning available jobs to machines via selecting from among a set of well-established dispatching rules.

{\bf Related work:} Industrial scheduling schemes for dynamic \gls{FJSPs} commonly employ heuristic dispatching rules, which define priority functions to assign available jobs to machines \cite{geiger2006rapid}. Typically, the scheduling process is decomposed into job sequencing and machine routing, and dispatching rules are designed for each of the two. A combined job sequencing and machine routing rule is then used as the overall dispatching strategy. Although dispatching rules are fast and react immediately to shop floor changes, they rules are usually myopic and highly dependent on the dataset structure. Moreover, no single dispatching rule performs best across all \gls{FJSPs} scenarios. Thus, more advanced scheduling approaches are required.


Another widely adopted method to handle scheduling with dynamic events is event-triggered rescheduling. In this paradigm, the scheduler monitors shop floor events such as job arrivals or machine breakdowns, and triggers a rescheduling procedure using metaheuristics or mathematical optimization based on the updated floor state. Metaheuristics are high-level search frameworks that guide subordinate heuristics to explore large combinatorial solution spaces efficiently, using stochastic or population-based mechanisms. For example, particle swarm optimization, a swarm-intelligence metaheuristic, was used to minimize energy consumption in \gls{FJSPs} environments with machine breakdowns \cite{nouiri2018towards}.  Genetic algorithms, an evolutionary metaheuristic, have been applied for rescheduling under dynamic disturbances \cite{zhang2021digital}. On the other hand, an iterative \gls{MILP} approach was proposed to solve identical parallel machine rescheduling problems with future job arrivals \cite{tighazoui2021predictive}. An event-triggered \gls{MILP} was used for rescheduling under machine breakdowns or raw material shortages while minimizing deviations from the original schedule \cite{figueroa2024adaptive}. A similar idea was applied to reschedule production lines to handle dynamic disturbances \cite{fu2025digital}. Although event-triggered rescheduling methods can often produce high-quality schedules, they are typically computationally expensive for real-time scheduling scenarios.



\gls{RL} has emerged as a promising direction for dynamic scheduling because it can balance scheduling quality with real-time decision making. In \gls{RL}, an agent continuously observes the shop floor state and takes scheduling actions to maximize a cumulative reward. Early attempts to apply \gls{RL} to scheduling problems date back to the 1990s \cite{riedmiller1999neural}. In the context of \gls{FJSPs}, classical tabular Q-learning was used to select dispatching rules in environments with random job arrivals \cite{aydin2000dynamic}, and a multi-agent tabular Q-learning framework was adopted to coordinate machine selection and job sequencing decisions \cite{bouazza2017distributed}. 
However, tabular \gls{RL} approaches rely on explicit lookup tables, which scale poorly as the state space grows, limiting their applicability in complex scheduling environments. To address these limitations, recent studies have explored \gls{DRL}. By using neural
networks for function approximation, \gls{DRL} can handle larger state spaces and capture complex features. For instance, \cite{waschneck2018optimization} utilized deep Q-learning to maximize machine utilization of a \gls{FJSPs} under random machine breakdowns; while \cite{zhang2023deep} leveraged an actor-critic network to solve \gls{FJSPs} with stochastic processing times to minimize makespan. 

Specifically for \gls{FJSPs} with random job arrivals, seven elaborately-designed state features were introduced and a two-hierarchy deep Q network was trained to select hand-designed composite rules at each rescheduling point \cite{luo2021dynamic}. Similarly, \cite{liu2022deep} categorized the $25$ abstract features into $6$ channels and developed a hierarchical \gls{DRL} framework with a double deep Q-network for job arrivals with constant rates. More recently, \gls{PPO} with an attention-based policy network was employed to directly assign pending jobs for the total tardiness minimization \cite{zhao2023drl}. A heterogeneous graph neural network is deployed to create embeddings for \gls{PPO} to solve \gls{FJSPs} with random job arrivals and limited transportation resources \cite{zhang2023dynamic}.

Despite these promising results, several limitations remain. First, some \gls{DRL} methods adopt complex state representations and sophisticated network architectures to encode shop floor information \cite{zhao2023drl,zhang2023dynamic}, which increases model complexity and training difficulty. Additionally, most approaches consider constant job arrival rates and assume homogeneous machines and jobs \cite{luo2021dynamic,liu2022deep,luo2020dynamic}.  Here, we provide an event-based \gls{DRL} framework, where the agent only applies actions at each event (i.e., operation done or new job arrived). The framework adopts concise state and action representations, and is trained with \gls{PPO} using lightweight \gls{MLP}s as the actor and critic networks. We benchmark our \gls{DRL} against an event-based online rescheduling method, which we call \gls{AT-MILP}, demonstrating the efficacy of the proposed \gls{DRL} framework for dynamic \gls{FJSPs}. The main contributions of this work are: 
\begin{itemize}
\item We propose an event-based \gls{DRL} environment for \gls{FJSPs} with random job arrivals aimed at minimizing makespan.



\item We modify the traditional \gls{MILP} method into an event-based method for dealing with dynamic job arrivals, and use it as a strong baseline for the proposed \gls{DRL}.

\item We study the performance of the proposed \gls{DRL} and \gls{AT-MILP} on datasets with varying heterogeneity, and show that our \gls{DRL} method achieves good performance, especially when the datasets are heterogeneous.
\end{itemize}

\section{Problem Formulation}
\label{sec:Problem def}
The \gls{FJSPs} we consider comprise a finite set of jobs 
$\mathcal{J}=\{J_1,\dots,J_n\}$ and a set of machines $\mathcal{M}=\{M_1,\dots,M_m\}$. 
Each job $J_i\in\mathcal{J}$ is composed of a sequence of operations 
$O_i=\{O_{i1},\dots,O_{in_i}\}$. The operation $O_{ij}$ of job $J_i$ can be processed on any
machine selected from a set of compatible machines $\mathcal{M}_{ij}\subseteq\mathcal{M}$. The processing time of operation $O_{ij}$ on a compatible machine $M_k \in \mathcal{M}_{ij}$ is denoted by $p_{ij}^k\in\mathbb{R}^+$, and can vary among different machines. Each job $J_i$ has a stochastic arrival time, $A_i$, which follows a probability distribution discussed in Section \ref{sec:simulation}. The goal of \gls{FJSPs} in this work is to minimize the final makespan 
$C_{\max}$. The scheduling and processing of operations in \gls{FJSPs} must satisfy the following constraints:


\begin{itemize}
  \item Each machine can process at most one operation at a time (no overlap constraint).
  \item The execution order of the set of operations within the same job between successive operations must be respected (precedence constraint). 
  \item No operation from any job has priority and each operation should be processed without interruption (non-preemption and no interruption constraint).
  \item The processing time of each operation on each machine is known and deterministic.
  \item Machine setup time and job materials transportation time are neglected. 
  \item Job arrivals $A_i$ are exogenous random variables revealed on-line.
\end{itemize}

The \gls{FJSPs} with a realization of job arrivals is mathematically formulated as:

\begin{subequations}\label{eq:fjsp_logic}
\small
\begin{align}
\min_{x,b,c}\ & C_{\max} \notag \\[2pt]
\text{s.t.}\ 
& \sum_{k\in\mathcal{M}_{ij}} x_{ij}^k = 1,
  \quad \forall i,j
  \label{eq:assign} \\[-2pt]
& c_{ij} = b_{ij} + \sum_{k\in\mathcal{M}_{ij}} x_{ij}^k p_{ij}^k,
  \quad \forall i,j
  \label{eq:comp} \\[-2pt]
& b_{i,j+1} \ge c_{ij},
  \quad \forall i,j
  \label{eq:prec} \\[-2pt]
& b_{i1} \ge A_i,
  \quad \forall i
  \label{eq:release} \\[-2pt]
& C_{\max} \ge c_{ij},
  \quad \forall i,j
  \label{eq:makespan} \\[4pt]
& \forall (i,j)\neq(p,q),\ 
  \forall k \in \mathcal{M}_{ij}\cap\mathcal{M}_{pq}: \notag \\
& (x_{ij}^k = 1 \wedge x_{pq}^k = 1)
  \Rightarrow
  \left(b_{pq} \ge c_{ij} \ \vee\  b_{ij} \ge c_{pq}\right)
\label{eq:machine} \\
& x_{ij}^k \in \{0,1\},\quad
  b_{ij},\ c_{ij} \ge 0 .
\end{align}
\end{subequations}

The objective is to minimize the makespan $C_{\max}$ subject to several constraints. 
Equation~\eqref{eq:assign} ensures that each operation is assigned to exactly one compatible machine, where $x_{ij}^k \in \{0,1\}$ is a binary decision variable indicating whether operation $O_{ij}$ is assigned to machine $M_k$. 
Equation~\eqref{eq:comp} defines the completion time $c_{ij}\in \mathbb{R}_{+}$ of each operation based on its start time $b_{ij}\in \mathbb{R}_{+}$ and processing time $p_{ij}^k\in \mathbb{R}_{+}$ on the assigned machine $M_k$. 
Constraint~\eqref{eq:prec} enforces the precedence constraint within each job, while~\eqref{eq:release} ensures that the first operation of each job does not start before its arrival time  $A_i \in \mathbb{R}_{+}$. 
Equation~\eqref{eq:makespan} defines the makespan as the maximum completion time across all operations. 
Constraint~\eqref{eq:machine} enforces machine capacity using a logical
disjunctive constraint. If operations $O_{ij}$ and $O_{pq}$ are both assigned to machine $M_k$,
then either $O_{ij}$ precedes $O_{pq}$ or vice versa.
For computational implementation in this work, the disjunctive constraint in~\eqref{eq:machine} is linearized using binary sequencing variables and standard big-$M$ constraints.

\section{Proposed methods for \gls{FJSPs} with random job arrivals}
\label{sec:our method}
A FJSP can be modeled as an \gls{MDP}, making it amenable to \gls{DRL} methods. 

\subsection{\gls{MDP} formulation of dynamic scheduling environment} \label{sec:MDP}
An \gls{MDP} model consists of a tuple $(\mathcal{S},\mathcal{A},P,r,\gamma)$
, where $\mathcal{S}$ is the state space, $\mathcal{A}$ the action space, $P(s' \vert s,a)$ the transition kernel, $r(s,a,s')$ the immediate reward, and $\gamma\in[0,1]$ the discount factor. The goal is to find a policy $\pi:\mathcal{S}\to\mathcal{A}$ maximizing the discounted expected return
\[
G(\pi)=\mathbb{E}_{\pi,P}\Big[\sum_{t=0}^{T-1} \gamma^t r(s_t,a_t,s_{t+1})\Big].
\]


We model the \gls{FJSPs} with random job arrivals as a discrete-event \gls{MDP}. The index $t \in \mathbb{N}$ denotes decision events, occurring at job arrivals or operation completions. The elapsed physical time between consecutive decision events is variable. 

\subsubsection{State}
\label{subsec:state}
A state 
\[
s_t = ( q_t,\; m_t,\; \tau_t\!) \;,
\]
where
\begin{itemize}
  \item $q_t$: a vector of size $|\mathcal{J}|$ indicating the next operation index of each job.
  \item $m_t$: a vector of size $|\mathcal{M}|$ indicating the next free time of each machine.
  \item $\tau_t$: a vector of size $|\mathcal{J}|$ indicating the next available time for each job. For any job $J_i$ that has not arrived, its state is set to a large value. 
\end{itemize}
Note that for \gls{FJSPs} with random job arrivals, the job state cannot be observed for any job that has not arrived, making the problem a \gls{POMDP} as described below in \ref{sec:framework RL}. 


\subsubsection{Action}
In the \gls{FJSPs} literature, two main approaches are used to define an action $a_t$:
\begin{itemize}
    \item directly selecting a compatible job--machine pair $(J_i, M_k)$;
    \item selecting from a pool of dispatching rules, either tailor-made or well-established.
\end{itemize} 
Directly selecting a $(J_i, M_k)$ can be optimal \cite{zhao2023drl, liu2025preference}, but the training complexity grows rapidly as the problem size increases. With dispatching rules as the action space, on the other hand, the training is much easier to converge as shown in Section \ref{sec:simulation}. Moreover, the action space remains interpretable and can be arbitrarily enlarged by appending additional rules, ameliorating the suboptimality issue. 

Dispatching rules determine scheduling decisions through two steps. 
First, a job sequencing rule prioritizes the next operation among the available jobs based on simple criteria such as processing time or arrival order. 
Second, a machine routing rule assigns the selected operation to one of its compatible machines, for example by choosing the machine with the earliest completion time. 
In this work, we adopt this rule-based approach and use five dispatching rules for job sequencing (\gls{SPT}, \gls{LPT}, \gls{FIFO}, \gls{LIFO}, \gls{MWR}), and two dispatching rules for machine routing (\gls{MIN}, \gls{MINC}). Therefore, the action space consists of $5 \times 2 = 10$ dispatching rule combinations.

\subsubsection{Transition kernel}
\label{subsec:transition}

The transition kernel is stochastic and is defined as
\[
P(s' \mid s,a) = \Pr(s_{t+1}=s' \mid s_t=s, a_t=a),
\]
where the probability distribution of $s_{t+1}$ is determined by the random job arrival process, given that the operation processing times are assumed to be deterministic.

\subsubsection{Expected return}
\label{subsec:reward}
 The objective of \gls{FJSPs} is to minimize the final makespan, which is naturally sparse (only obtainable at the end of each episode). We adopt the reward shaping strategy that defines the reward $r(s_t,a_t,s_{t+1})=C(s_t)-C(s_{t+1})$ as the difference between the current makespan of the (partial) schedule at $s_t$ and $s_{t+1}$. We adopt the episodic setting where an episode terminates when all operations are finished. We set the discount factor $\gamma=1$, and the expected return in a single episode is 
\[
\begin{aligned}
G(\pi) &= \mathbb{E}_{\pi,P}\Big[\sum_{t=0}^{T-1} \gamma^t r(s_t,a_t,s_{t+1})\Big] \\
       &= \sum_{t=0}^{T-1} \big(C(s_t)-C(s_{t+1})\big) \\
       &= C(s_0)-C(s_T) \\
       &= -C(s_T).
\end{aligned}
\]
 
 Because we always assume that we start with an empty schedule, maximizing the expected return under this reward shaping is equivalent to minimizing the final makespan.

\subsection{\gls{PPO} algorithm}
\label{sec:PPO}
The actor-critic \gls{PPO} ~\cite{schulman2017ppo} optimizes a stochastic policy $\pi_\theta$ via actor while estimating a critic value function $V_\phi$. The probability of sampling an action $a_t$ given the current state $s_t$ from the policy $\pi_\theta$ is denoted by $\pi_\theta(a_t \mid s_t)$. \gls{PPO} uses a \emph{surrogate} objective to prevent large policy updates; one common objective is the \emph{clipped loss}:
\[
L^{\text{CLIP}}(\theta)
= \mathbb{E}_{t}\Big[
\min\!\big( 
\rho_t(\theta)\,\hat{A}_t,\;
\operatorname{clip}(\rho_t(\theta),1-\epsilon,1+\epsilon)\,\hat{A}_t
\big)
\Big],
\]
where
\[
\rho_t(\theta) = \frac{\pi_\theta(a_t\mid s_t)}{\pi_{\theta_{\text{old}}}(a_t\mid s_t)}
\]
is the importance ratio between the updated policy $\pi_\theta$ and the behavior policy $\pi_{\theta_{\text{old}}}$ used for collecting trajectories. After the rollout, the parameters $\theta$ are updated 
for several epochs while keeping $\theta_{\text{old}}$ fixed. The clipping operator restricts $\rho_t(\theta)$ to the interval $[1-\epsilon,1+\epsilon]$, preventing overly large policy updates and improving training stability. The advantage estimate $\hat{A}_t$ is computed using $V_\phi$, where $\phi$ is updated by minimizing a squared value error. In this work, we employ \emph{Generalized Advantage Estimation} (GAE) \cite{schulman2015high}, which estimates the advantage from temporal-difference residuals
\[
\delta_t = r_t + \gamma V_\phi(s_{t+1}) - V_\phi(s_t),
\]
and aggregates them using a smoothing parameter $\lambda \in [0,1]$ to balance bias and variance.

\subsection{Framework of the event-based \gls{DRL}}
\label{sec:framework RL}
\begin{figure}[!t]
    \centering
\includegraphics[width=\columnwidth]{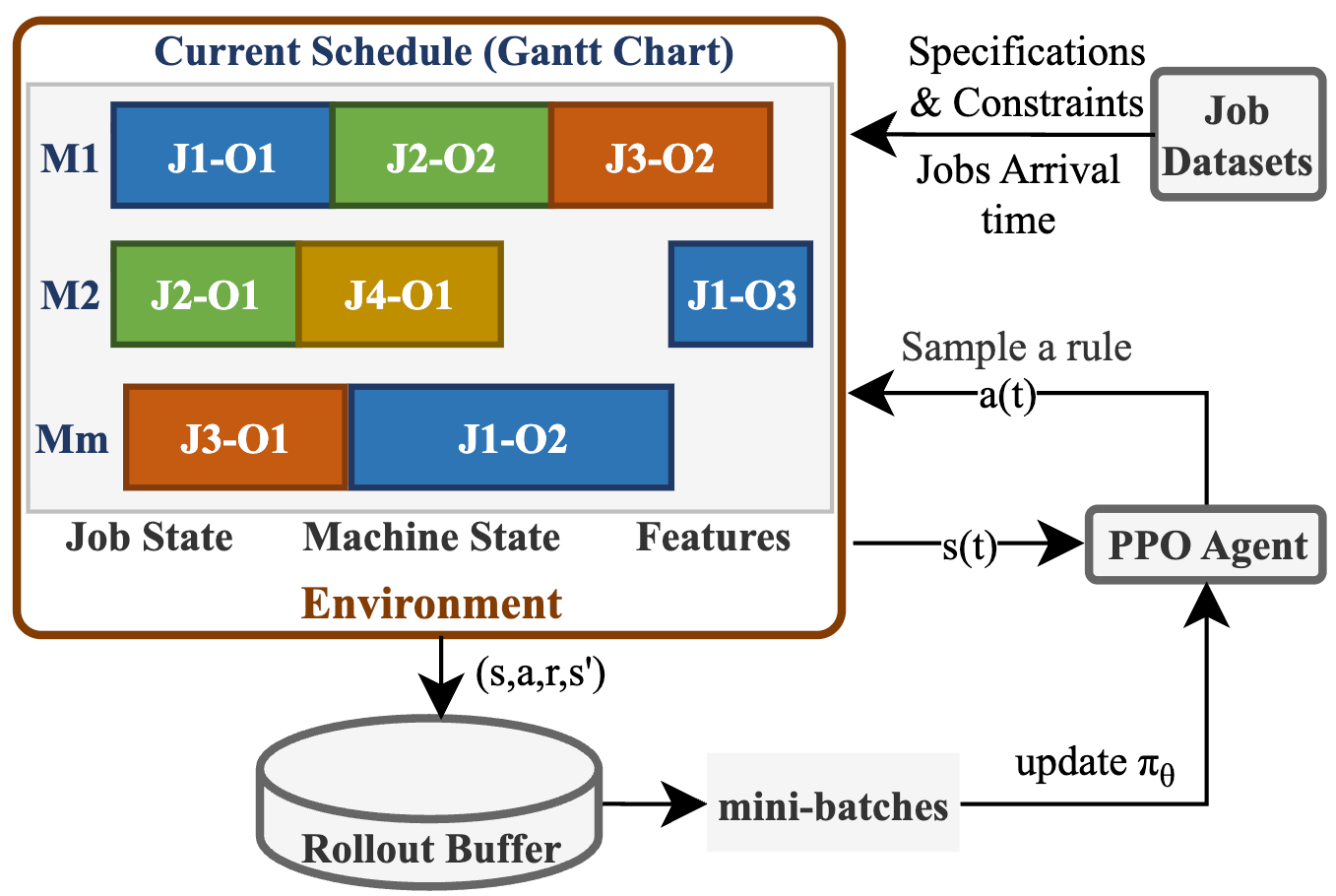}
    {\centering \caption{The framework of our \gls{DRL} for \gls{FJSPs} with random job arrivals.}\par     \label{fig:framework}}
    
\end{figure}

The overall framework of our method is depicted in Figure \ref{fig:framework}. This framework builds the \gls{FJSPs} environment by first taking the specifications (e.g., processing time of each job order, job-machine compatibility, sampled job arrival time) and the required constraints (e.g., job order precedence, no overlap) from a job shop scheduling dataset. It then defines the state observation, action space, reward, and action transition. The composed MDP trajectories are stored in a rollout buffer, which feeds mini-batches to train the \gls{PPO} agent. The \gls{PPO} agent employs lightweight \gls{MLP}s for both the actor and critic networks. Each \gls{MLP} comprises  two hidden layers with $256$ and $128$ neurons, respectively, and uses the Rectified Linear Unit (ReLU) activation function after each hidden layer. Based on the state observations, the \gls{PPO} agent outputs action logits through the policy network. 
A softmax function is applied to convert the logits into a probability distribution over dispatching rules, from which the agent samples an action that implicitly places an operation \textit{block} onto the current schedule Gantt chart.

A key design choice is the observation space of the \gls{POMDP}. The observation space includes the job state and machine state as discussed in Section \ref{subsec:state}. The agent observes the job state vector of the first $l$ arrived and unfinished jobs. If fewer than $l$ jobs have arrived, the remaining entries are padded by setting the job available time to $1$, corresponding to a large value after normalization. The hyperparameter $l$ is fixed, which ensures a constant input dimension for the \gls{MLP}s. We set $l=40$ as a sufficiently large value. Additionally, the observation space includes the average processing time of each job’s next operation, the current makespan, and the current \textit{event time} to provide useful global context for scheduling.




We model the \gls{FJSPs} as schedule building problems where the agent takes action at each \emph{event}. An event is defined as an operation done or a new job arrived. A critical parameter that controls when the agent sees the new job arrival and when the next operation can be placed is the \emph{event time} of the environment. The event time at each action step is the physical time of the earliest event on the current schedule. The event time is updated after each agent action, and it can remain unchanged if the earliest available machine is not scheduled at this action. In case there is no available operation to schedule at the current event time, the environment advances to the next event time.

\section{Simulations} 
\label{sec:simulation}
The proposed event-based framework is implemented with Python 3.9 on a PC with Apple M3 Pro 12 Cores and 36 GB RAM. The \gls{PPO} algorithm is implemented with \gls{SB3} \cite{raffin2021stable}. In this section, we present the results of two different datasets with respect to the job types and machine types: homogeneous datasets and heterogeneous datasets. The homogeneous datasets used in the simulation are generated similarly to \cite{luo2020dynamic}. Each job comprises a random number of operations sampled from a discrete uniform distribution $\mathrm{Unif}\{1,\ldots,4\}$. For any operation $O_{ij}$ and any compatible machine $M_k$, the processing time is drawn from a continuous uniform distribution $\mathrm{Unif}[1,100]$, and the number of compatible machines is sampled from a discrete uniform distribution $\mathrm{Unif}\{1,\ldots,M\}$. The shop initially contains $n_{\mathrm{ini}}$ jobs and subsequently receives $n_{\mathrm{add}}$ new jobs. These arrivals follow a Poisson process, and therefore the inter-arrival time $t_i$ between two successive jobs follows an exponential distribution with rate $\lambda$ (hence $E[t_i]=\frac{1}{\lambda}$ and $Var[t_i]=\frac{1}{\lambda^2})$. 

\begin{figure}[htbp]
  \centering
  \includegraphics[width=\linewidth]{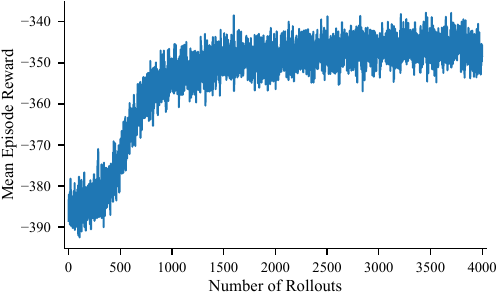}
  \caption{Mean episode cumulative reward as a function of the rollout number during training. The figure shows a learning curve for a homogeneous instance with $M=6$, $n_{\mathrm{ini}}=5$, $n_{\mathrm{add}}=15$, and $\lambda=0.2$.}
  \label{fig:training}
\end{figure}

To emulate realistic heterogeneity, in the heterogeneous datasets, jobs are classified as \emph{short} (85\%) or \emph{long} (15\%). Short jobs use base processing times $\mathrm{Unif}[10,30]$, while long jobs use $\mathrm{Unif}[75,300]$. Machines are divided into one \emph{fast} machine and the remaining \emph{slow} machines. When an operation is assigned to a slow machine, its processing time is multiplied by a slowness penalty $f$. We also enforce several operations to be only compatible with one machine, to encode the bottleneck machine. Similarly to homogeneous datasets, job arrivals follow a Poisson process. Full configuration details are in Table~\ref{tab:exp_parameters}.
The \gls{PPO} agent is trained and evaluated across the configurations listed in Table~\ref{tab:exp_parameters}, with the training hyperparameter values in Table~\ref{tab:train_parameters}. A systematic grid search of the hyperparameter space was not performed due to high computational cost. Figure~\ref{fig:training} shows a representative learning curve for a homogeneous instance with $M=6$, $n_{\mathrm{ini}}=5$, $n_{\mathrm{add}}=15$, and $\lambda=0.2$. The mean episode reward increases steadily and stabilizes after approximately $3000$ rollouts, indicating stable and effective training under the selected settings.


\begin{table}[htbp]
\centering
\caption{Parameter settings of job shop configuration used for \gls{DRL} training.}
\label{tab:exp_parameters}

\begin{subtable}{\columnwidth}
\centering
\caption{Homogeneous datasets \cite{luo2020dynamic}}
\begin{tabular}{p{0.55\columnwidth} p{0.36\columnwidth}}
\hline
\textbf{Parameter} & \textbf{Value} \\ \hline
Number of machines ($M$) & $6$ \\
Number of available machines of each operation & $\mathrm{Unif}\{1,\ldots,M\}$ \\
Number of initial jobs ($n_{\mathrm{ini}}$) & $\{5,15\}$ \\
Number of future jobs ($n_{\mathrm{add}}$) & $\{15,25\}$ \\
Number of operations belonging to a job & $\mathrm{Unif}\{1,\ldots,4\}$\\
Processing time of an operation on an available machine & $\mathrm{Unif}[1,100]$ \\
Rate parameter of the arrival exponential distribution ($\lambda$) & $\{0.2,0.05,0.02\}$ \\ \hline
\end{tabular}%
\end{subtable}

\vspace{0.5em}


\begin{subtable}{\columnwidth}
\centering
\caption{Heterogeneous datasets}
\label{tab:hetero}
\begin{tabular}{p{0.55\columnwidth} p{0.36\columnwidth}}
\hline
\textbf{Parameter} & \textbf{Value} \\ \hline
Number of machines ($M$) & $6$ \\
Number of fast machines ($M_{\mathrm{fast}}$) & $1$ \\
Number of available machines of each operation (except the bottleneck operations) & $\mathrm{Unif}\{1,\ldots,M\}$\\
Number of initial jobs ($n_{\mathrm{ini}}$) & $5$ \\
Number of future jobs ($n_{\mathrm{add}}$) & $15$ \\
Machine slowness penalty ($f$) & $3$  \\
Ratio of job types (short / long) & $0.85 \;/\; 0.15$ \\
Number of operations per job & $\mathrm{Unif}\{1,\ldots,4\}$\\
Base processing time per operation (short / long) & $\mathrm{Unif}[10,30] \;/\; \mathrm{Unif}[75,300]$ \\
Rate parameter of the arrival exponential distribution ($\lambda$) & $\{0.2,\;0.05,\;0.02\}$ \\ \hline
\end{tabular}
\end{subtable}

\end{table}


\begin{table}[htbp]
\centering
\caption{Hyperparameter values for
the \gls{PPO} training. The values are the same across the homogeneous and the heterogeneous datasets.} 
\label{tab:train_parameters}
\begin{tabular}{p{0.55\columnwidth} p{0.36\columnwidth}}
\hline
\textbf{Hyperparameter} & \textbf{Value} \\ \hline
Number of training steps & $1e7$ \\
Entropy coefficient & $0.02$ \\
Training epochs for updating network & $3$ \\
Clipping threshold $\epsilon$ & $0.2$ \\
Learning rate & $0.0001$ \\
Rollout size & $256$ \\
Minibatch size & $64$ \\
Discount factor $\gamma$ & $1$ \\
Smoothing factor $\lambda$ & $0.95$ \\ 
Value function coefficient & $0.5$ \\ 
\hline
\end{tabular}
\end{table}


\subsection{Baseline methods and datasets}
\label{sec:baseline methods}
Ideally, a scheduling policy would be evaluated against the optimal schedule obtained with full knowledge of all job arrivals in hindsight by solving a single \gls{MILP} over the entire planning horizon. However, computing such a clairvoyant benchmark is computationally infeasible for the problem sizes considered in this study. Therefore, we resort to two practical baselines: Arrival Triggered Mixed Integer Linear Programming (\gls{AT-MILP}) and \gls{Best HH}. 

The \gls{AT-MILP} computes an initial \gls{MILP} solution based on the initial available jobs on the floor, and then triggers a new \gls{MILP} problem whenever a new job arrives. At the moment of any new job arrival, the finished and in-progress operations from the previous solves cannot be changed, while only the remaining and newly arrived operations can be scheduled at each rescheduling. The objective of every \gls{AT-MILP} solution is to minimize the makespan based on the current jobs on the shop floor. The solutions of the \gls{AT-MILP} are generated by the Gurobi solver \cite{gurobi}, and the solver optimization time of each rescheduling is limited to $60$ seconds for a balance between scheduling performance and online applicability. 

On the other hand, the \gls{Best HH} is obtained by applying all 
10 dispatching rules to each test instance and selecting the one with the smallest final makespan. The winning rule may vary across shop floors and job arrival realizations, and hence cannot be known in advance. The \gls{Best HH} is at least as good as any single dispatching rule, and strictly better unless one rule already achieves the smallest makespan in every instance. To show that the \gls{Best HH} is a strong baseline, we also present evaluation results for individual dispatching rules that each perform best in at least some test instances: \gls{FIFO}+\gls{MINC} (Rule 1), \gls{LIFO}+\gls{MINC} (Rule 2), \gls{LPT}+\gls{MIN} (Rule 3). Detailed results are given in the following section.

\subsection{Scheduling performance} \label{sec:performance}
\subsubsection{Homogenous datasets}
We evaluate the proposed \gls{DRL} agent under multiple job shop configurations with different job arrival rate parameters. For each configuration, the dataset specifications remain identical to those used during training, while the job arrival realizations (i.e., arrival sequences and times) are resampled independently. Table~\ref{tab:regret_comparison} shows the average and the standard deviation of the results for five independent runs of each experiment. The \gls{DRL} policy consistently outperforms or matches the \gls{Best HH} across all problem settings. This indicates that the learned policy can identify dispatching decisions comparable to the best dispatching rule in hindsight. Furthermore, the \gls{AT-MILP} achieves the smallest makespan in every testing scenario, even with a short time limit at each optimization ($60$ seconds). A possible explanation is that the considered shop floors have homogeneous jobs and machines (i.e., sampled from identical distributions) and a constant job arrival rate. Under such conditions, the local optimum (or feasible solution) obtained by \gls{AT-MILP} at each decision step is already a near-optimal solution.


Moreover, when the arrival rate $\lambda = 0.02$, all methods exhibit similar performance for both problem scales. In this case, the expected inter-arrival time between successive jobs is $\frac{1}{\lambda} = 50$, which is comparable to the average job processing length. Consequently, many previously released jobs are nearly completed before new jobs arrive, significantly reducing scheduling conflicts and making the overall \gls{FJSPs} easier to solve.

From a practical perspective, the \gls{AT-MILP} approach is suitable when the job shop environment and arrival patterns are relatively homogeneous and the online decision time budget is sufficiently flexible (e.g., when the inter-arrival time between successive jobs exceeds the optimization time limit). Furthermore, when job arrivals are infrequent relative to the average job duration, simple dispatching rules can already achieve competitive performance.

\begin{table}[htbp]
\centering
\caption{Rounded mean makespan and the standard deviation of different methods under different problem sizes and varying job arrival rates $\lambda$. Each scenario is tested over 5 independent runs, and the best result is in bold font.}
\label{tab:regret_comparison}

\begin{subtable}[t]{\columnwidth}
\centering
\caption{Small-scale problem (6 machines, 20 jobs)}
\begin{tabular}{lccc}
\toprule
\textbf{Methods} & $\lambda=0.2$ & $\lambda=0.05$ & $\lambda=0.02$ \\
\midrule
\gls{AT-MILP} & \textbf{307 $\pm$ 14} & \textbf{416 $\pm$ 35} & \textbf{763 $\pm$ 174} \\
\gls{DRL} (Ours) & 338 $\pm$ 18  & 420 $\pm$ 36 & 764 $\pm$ 173 \\
\gls{Best HH} & 378 $\pm$ 15 & 421 $\pm$ 42 & 765 $\pm$ 173 \\  
Rule 1 & 396 $\pm$ 22 & 428 $\pm$ 43 & 765 $\pm$ 173 \\ 
Rule 2 & 419 $\pm$ 15 & 429 $\pm$ 44 & 765 $\pm$ 173 \\ 
Rule 3 & 409 $\pm$ 32 & 434 $\pm$ 30 & 770 $\pm$ 171 \\ 
\bottomrule
\end{tabular}
\end{subtable}

\vspace{1ex}

\begin{subtable}[t]{\columnwidth}
\centering
\caption{Large-scale problem (6 machines, 40 jobs)}
\begin{tabular}{lccc}
\toprule
\textbf{Methods} & $\lambda=0.2$ & $\lambda=0.05$ & $\lambda=0.02$ \\
\midrule
\gls{AT-MILP} & \textbf{560 $\pm$ 14} & \textbf{596 $\pm$ 28} & \textbf{1380 $\pm$ 175} \\
\gls{DRL} (Ours) & 595 $\pm$ 23 & 627 $\pm$ 39 & \textbf{1380 $\pm$ 175} \\
\gls{Best HH} & 619 $\pm$ 16 & 634 $\pm$ 15 & \textbf{1380 $\pm$ 175} \\
Rule 1 & 658 $\pm$ 33 &   684 $\pm$ 44 &   \textbf{1380 $\pm$ 175} \\
Rule 2 & 657 $\pm$ 34 &   652 $\pm$ 11 &   \textbf{1380 $\pm$ 175} \\
Rule 3 & 644 $\pm$ 33 &   705 $\pm$ 58 &   \textbf{1380 $\pm$ 175} \\
\bottomrule
\end{tabular}
\end{subtable}

\end{table}

\subsubsection{Heterogeneous datasets}
Next, we train and evaluate our \gls{DRL} on heterogeneous datasets described in Table \ref{tab:exp_parameters} with a shop floor configuration of $6$ machines, $5$ initial jobs, and $15$ upcoming jobs, with different job arrival rates. Each rate parameter is evaluated over $15$ independent runs. The resulting average makespan and its standard deviation are reported in Table~\ref{tab:regret_heter} for the different arrival rates. The results show that this time, the \gls{AT-MILP} performs significantly worse, particularly for higher job arrival rates $\lambda$. This performance degradation can be attributed to the myopic nature of the \gls{AT-MILP} when facing heterogeneous shop floors combined with stochastic job arrivals. In such cases, the \gls{AT-MILP} may already occupy fast machines with relatively short jobs when longer and critical jobs arrive, leading to suboptimal scheduling decisions. In contrast, the proposed \gls{DRL} approach learns policies that better adapt to heterogeneous
shop floor conditions. 

\begin{figure}[htbp]
  \centering
  \includegraphics[width=\linewidth]{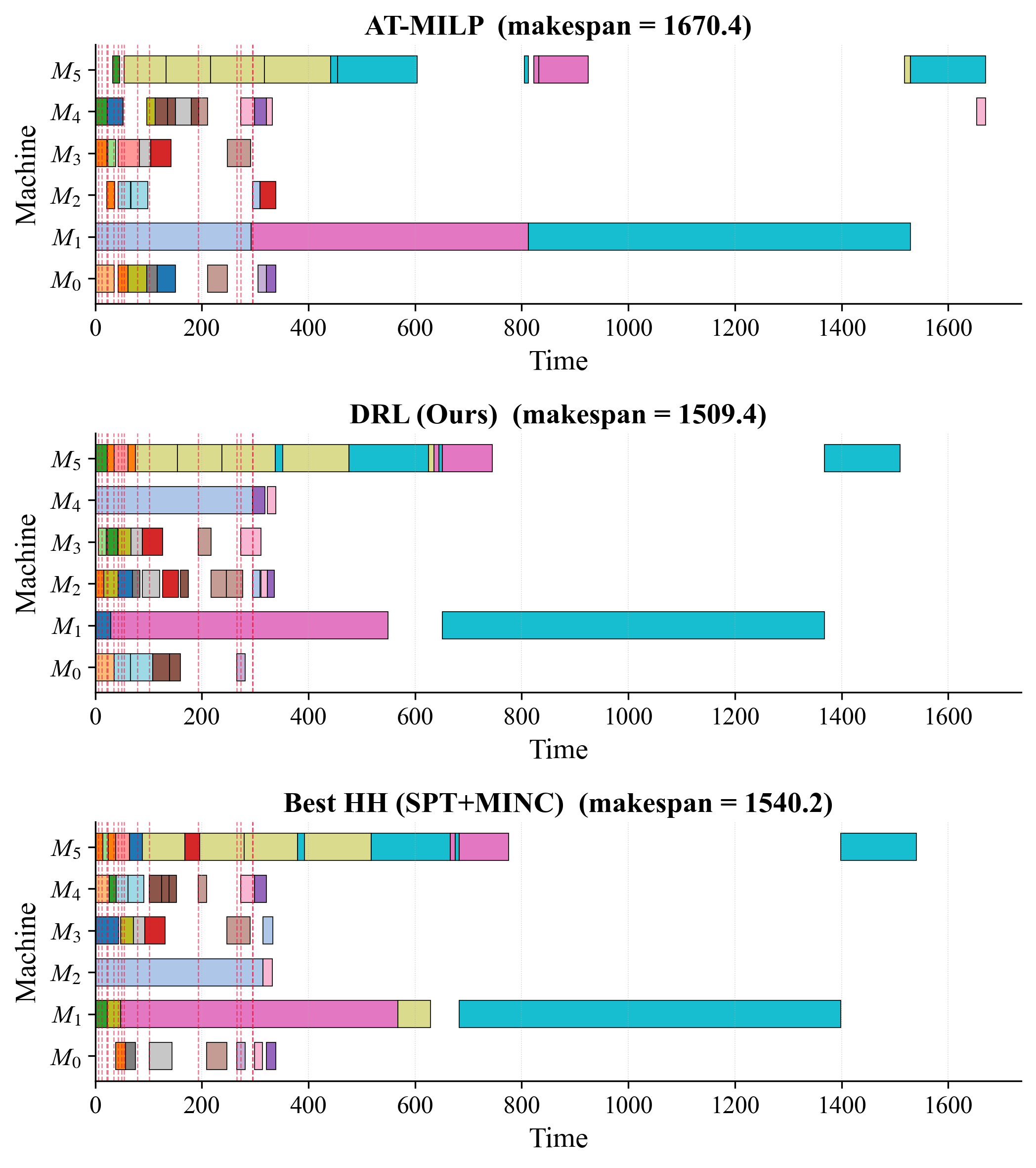}
  \caption{Schedules of three methods on one heterogeneous case with $M=6$, $n_{\mathrm{ini}}=5$, $n_{\mathrm{add}}=15$, and $\lambda=0.05$. The \gls{Best HH} in this run is \gls{SPT} for job sequencing and \gls{MINC} for machine routing. The vertical dashed lines in red indicate the incoming job arrival times. Jobs are represented with different colors on the Gantt chart, and the blocks with the same color represent different operations of the same job.}
  \label{fig:gantt}
\end{figure}

To further illustrate this behavior, the schedules from all methods for one test run are visualized using Gantt charts in Figure~\ref{fig:gantt}. In this example, machine $M_5$ is the only fast machine, while several operations are exclusively compatible with the bottleneck machine $M_1$. Under this configuration, the proposed \gls{DRL} achieves the smallest makespan, whereas the \gls{AT-MILP} struggles due to its myopic decision-making at individual arrival events. 



\begin{table}[htbp]
\centering
\caption{Rounded mean makespan and standard deviation of different methods under the shop floor of 6 machines and 20 jobs with varying job arrival rates $\lambda$. Each scenario is tested over $15$ independent runs, and the best result is in bold font.}
\label{tab:regret_heter}

\begin{tabular}{lccc}
\toprule
\textbf{Methods} & $\lambda=0.2$ & $\lambda=0.05$ & $\lambda=0.02$ \\
\midrule
\gls{AT-MILP}& 1642 $\pm$ 14 & 1660 $\pm$ 19 & 1757 $\pm$ 109 \\
\gls{DRL} (Ours) & \textbf{1478 $\pm$ 52}  & \textbf{1550 $\pm$ 86} & \textbf{1753 $\pm$ 174} \\
\gls{Best HH} & 1499 $\pm$ 83 & 1564 $\pm$ 104 & 1797 $\pm$ 197 \\
Rule 1 & 1558 $\pm$ 107 &  1620$\pm$ 140 & 1817 $\pm$ 201 \\ 
Rule 2 & 1535 $\pm$ 102 &  1623$\pm$125  & 1820 $\pm$ 211 \\ 
Rule 3 & 2014 $\pm$ 50 &  2029$\pm$59  & 2120 $\pm$ 92 \\ 
\bottomrule
\end{tabular}
\end{table}





\section{Conclusion}
\label{sec:conclu}
We develop an event-based \gls{DRL} framework to tackle dynamic job arrivals in \gls{FJSPs}. We design the action space as a set of well-established dispatching rules; the action space can be arbitrarily enlarged by appending other dispatching rules. We employ the actor-critic \gls{PPO} algorithm with lightweight \gls{MLP}s for policy training, making our framework readily interpretable and adaptable. The simulation results show that the scheduling performance of our \gls{DRL} beats or ties with the best heuristic dispatching rule in hindsight. Moreover, we benchmark against a method called \gls{AT-MILP}, which resolves online a new \gls{MILP} problem when jobs arrive. In practice, the employment of this \gls{AT-MILP} needs to be decided carefully. When the datasets are homogeneous and the online solution time limit is not tight, using \gls{AT-MILP} is favored; otherwise, deploying our \gls{DRL} improves performance. An interesting direction for future research is to investigate different job arrival patterns and additional types of heterogeneous datasets to enhance the robustness of this framework.



\bibliographystyle{IEEEtran}
\bibliography{ref}             
\end{document}